\newcommand{\xbe}{\in}

\newcommand{\xbl}{\lambda}
\newcommand{\xbm}{\mu}

\newcommand{\xcc}{\subseteq}

\newcommand{\xcj}{\Leftrightarrow}

\newcommand{\xcl}{\vdash}
\newcommand{\xcm}{\models}
\newcommand{\xcn}{\hspace{0.2em}\sim\hspace{-0.9em}\mid\hspace{0.58em}}

\newcommand{\xcp}{\rightarrow}

\newcommand{\xcs}{\cap}

\newcommand{\xcv}{\cup}

\newcommand{\xDH}{\item }

\newcommand{\xEI}{\begin{itemize}}
\newcommand{\xEJ}{\end{itemize}}

\newcommand{\xEh}{\begin{enumerate}}

\newcommand{\xEj}{\end{enumerate}}

\newcommand{\xEn}{\begin{description}}
\newcommand{\xEp}{\end{description}}

\newcommand{\Xl}{\ldots}

\newcommand{\ol}{\overline}

\newcommand{\bl}{\begin{lemma} \rm}
\newcommand{\el}{\end{lemma}}
\newcommand{\br}{\begin{remark} \rm}
\newcommand{\er}{\end{remark}}
\newcommand{\be}{\begin{example} \rm}
\newcommand{\ee}{\end{example}}
\newcommand{\bco}{\begin{corollary} \rm}
\newcommand{\eco}{\end{corollary}}
\newcommand{\bc}{\begin{claim} \rm}
\newcommand{\ec}{\end{claim}}
\newcommand{\bfa}{\begin{fact} \rm}
\newcommand{\efa}{\end{fact}}
\newcommand{\bp}{\begin{proposition} \rm}
\newcommand{\ep}{\end{proposition}}
\newcommand{\bd}{\begin{definition} \rm}
\newcommand{\ed}{\end{definition}}
\newcommand{\bcs}{\begin{construction} \rm}
\newcommand{\ecs}{\end{construction}}
\newcommand{\bcd}{\begin{condition} \rm}
\newcommand{\ecd}{\end{condition}}
\newcommand{\bt}{\begin{theorem} \rm}
\newcommand{\et}{\end{theorem}}
\newcommand{\bn}{\begin{notation} \rm}
\newcommand{\en}{\end{notation}}
\newcommand{\bfi}{\begin{bild} \rm}
\newcommand{\efi}{\end{bild}}
\newcommand{\bsta}{\begin{statement} \rm}
\newcommand{\esta}{\end{statement}}
\newcommand{\bcom}{\begin{comment} \rm}
\newcommand{\ecom}{\end{comment}}
\newcommand{\bdia}{\begin{diagram} \rm}
\newcommand{\edia}{\end{diagram}}

\newcommand{\bfc}{\begin{figure}[htb] \begin{center}}
\newcommand{\efc}{\end{center} \end{figure}}

\documentclass{article}

\usepackage{amssymb,latexsym,epic,eepic,rotating}

\title{KI, Philosophie, Logik
\thanks{File: Ein - [Sch18j]
}
}

\author{Karl Schlechta
\thanks{
schcsg@gmail.com - https://sites.google.com/site/schlechtakarl/ -
Koppeweg 24, D-97833 Frammersbach, Germany}
\thanks{
Retired, formerly: Aix-Marseille Universit\'{e}, CNRS, LIF UMR 7279, F-13000
Marseille, France
}
}

\begin{document}

\newtheorem{lemma}{Lemma}[section]
\newtheorem{theorem}[lemma]{Theorem}
\newtheorem{proposition}[lemma]{Proposition}
\newtheorem{corollary}[lemma]{Corollary}
\newtheorem{claim}[lemma]{Claim}
\newtheorem{fact}[lemma]{Fact}
\newtheorem{remark}[lemma]{Remark}
\newtheorem{definition}{Definition}[section]
\newtheorem{construction}{Construction}[section]
\newtheorem{condition}{Condition}[section]
\newtheorem{example}{Example}[section]
\newtheorem{notation}{Notation}[section]
\newtheorem{bild}{Figure}[section]
\newtheorem{comment}{Comment}[section]
\newtheorem{statement}{Statement}[section]
\newtheorem{diagram}{Diagram}[section]

\renewcommand{\labelenumi}
  {(\arabic{enumi})}
\renewcommand{\labelenumii}
  {(\arabic{enumi}.\arabic{enumii})}
\renewcommand{\labelenumiii}
  {(\arabic{enumi}.\arabic{enumii}.\arabic{enumiii})}
\renewcommand{\labelenumiv}
  {(\arabic{enumi}.\arabic{enumii}.\arabic{enumiii}.\arabic{enumiv})}

\maketitle

\setcounter{secnumdepth}{3}
\setcounter{tocdepth}{3}

\begin{abstract}

This is a short (and personal) introduction in German to the connections between
artificial intelligence, philosophy, and logic, and to the author's work.

Dies ist eine kurze (und pers\"onliche) Einf\"uhrung in die Zusammenh\"ange
zwischen K\"unstlicher Intelligenz, Philosophie, und Logik, und in die
Arbeiten des Autors.

\end{abstract}

\tableofcontents
\clearpage

% *** BEGIN LATEX SOURCE ein-1.tex ***
%
% Aus Karltex File: ein-1.ms
%
%
% ae, oe kann man systematisch aendern, nur bei ue aufpassen
% Grosse Ae machen
\section{
KI und Philosophie - Beispiel Normalfall und Deontische Logik
}
\subsection{
Ein Beispiel aus der KI (= K\"unstliche Intelligenz)
}

V\"ogel fliegen - gemeint ist: k\"onnen fliegen.

Nicht alle V\"ogel k\"onnen fliegen, Pinguine k\"onnen nicht fliegen,
Brath\"ahnchen
auch nicht  \Xl. Aber wenn wir an einen Vogel denken, denken wir daran,
dass er
wohl fliegen kann.

Wir k\"onnen das so ausdr\"ucken: ``Normale'' V\"ogel k\"onnen fliegen.

Wenn man einen K\"afig f\"ur V\"ogel bauen will, wird man an ``normale''
V\"ogel denken, und ein Dach planen. Wenn man aber weiss, dass es sich
um einen K\"afig f\"ur Pinguine handelt, wird man auf das Dach verzichten.

Da es oft sehr viele Ausnahmen gibt, an die man nicht unbedingt alle
denkt (und die man nicht alle programmieren will),
fasst man die \"ublichen F\"alle eben mit dem Wort ``normal''
zusammen, und die Ausnahmen sind dann ``nicht normal''.
Erst wenn man Gr\"unde hat, anzunehmen, dass ein Vogel kein ``normaler''
Vogel ist, wird man an die Ausnahmen denken. Das muss aber explizit
angegeben sein.

Es ist einfacher,
Regeln f\"ur die ``normalen'' F\"alle zu programmieren, und Ausnahmen explizit
anzugeben, als von vornherein an alle Ausnahmen zu denken, und sie
zu programmieren.

Daher kommt das Interesse der KI f\"ur die Beschreibung der ``normalen''
F\"alle, die in Regeln mit Ausnahmen resultiert.

Man kann die F\"alle jetzt nach ihrer ``Normalit\"at'' oder aus historischen
Gr\"unden umgekehrt ``Abnormalit\"at'' ordnen, $Fall_{1}<Fall_{2}$ dr\"uckt
aus,
dass $Fall_{1}$ weniger abnormal ist, als $Fall_{2},$ oder, dass
$Fall_{1}$ normaler ist als $Fall_{2}.$

Ein Fall minimaler Abnormalit\"at oder maximaler Normalit\"at
ist also minimal bez\"uglich $<,$ das heisst, es gibt keinen Fall,
der bez\"uglich $<$ kleiner ist.

Man interessiert sich f\"ur diese minimalen Punkte - genauer: Modelle,
siehe Section \ref{Section Mini} (page \pageref{Section Mini})
\subsection{
Philosophische Logiken
}

Philosophische Logiken besch\"aftigen sich z. B. mit Begriffen wie
``m\"oglich'',
(moralisch) ``geboten'', etc., und versuchen, diese Begriffe zu analysieren,
darum nennt man diesen (und \"ahnliche) Bereich der Philosophie auch
``analytische Philosphie''.

Die Logik, die sich mit dem besch\"aftigt, was (moralisch) geboten ist,
nennt man
``deontische Logik'' (das Wort kommt aus dem Griechischen).

Die moralisch besten Situationen sind dann die, in denen alle
Gebote befolgt sind, also zB niemand ermordet wird, keine Sachen
gestohlen werden, etc.

Man kann wieder die Situationen ordnen, zB bedeutet
$Situation_{1}<Situation_{2}$
dass in $Situation_{1}$ weniger Gebote verletzt werden als in
$Situation_{2},$
und eine perfekte Situation ist dann minimal bez\"uglich $<.$

Hier ist dann eine Parallele zwischen KI und Philosophie: man
interessiert sich f\"ur die F\"alle, die bez\"uglich einer
gewissen, abstrakten, Ordnung $<,$ minimal sind - die ``F\"alle''
sind dann wieder Modelle, die in der Philosophie oft
``m\"ogliche Welten'' genannt werden.

(Der grosse amerikanische Philosoph David Lewis hat geglaubt, dass alle
m\"oglichen Welten wirklich existieren - was ich f\"ur Unsinn halte  \Xl.)
\section{
Logik - Grundbegriffe
}
\subsection{
Herleitung und Axiome
}
\subsubsection{
Ziel der Logik
}

Das urspr\"ungliche Ziel der Logik ist, absolut gesicherte Erkenntnisse
zu erreichen.

Da sich das Ziel als zu ehrgeizig herausgestellt hat, hat man es im Laufe
der Jahrtausende relativiert:
 \xEh
 \xDH
Von absolut sicheren Erkenntnissen ist man zu sicheren Erkenntnissen
unter genau beschriebenen, expliziten Voraussetzungen (Axiomen)
\"ubergegangen.
 \xDH
W\"ahrend die Axiome urspr\"unglich als ``selbverst\"andlich'',
``unmittelbar einsichtig'' etc. angenommen wurden, ist man zu
irgendwelchen Annahmen \"ubergegangen, deren G\"ultigkeit nicht weiter
hinterfragt wird, man nimmt sie halt an.
 \xDH
Man pr\"azisiert jetzt die Logik, mit deren Hilfe man etwas schliesst,
verschiedene Bereiche haben ihre verschiedenen Logiken, mit denen man
\"uber Eigenschaften in diesen Bereichen schliessen kann, zB \"uber
``m\"oglich'', ``geboten'', ``normalerweise'', etc.
 \xEj
\subsubsection{
Sprache
}

Allen diesen Logiken ist gemeinsam, dass man erst die Sprache, in der die
Aussagen formuliert werden, pr\"azisiert. Das sind (meist sehr einfache)
Kunstsprachen, da sich die nat\"urlichen Sprachen als zu komplex
und vage herausgestellt haben.

Zum Beispiel gibt man sich ein Alphabet wie $p,q,r, \Xl.$ vor, und
Operatoren
wie ``nicht'', ``und'', ``oder'', mit denen man aus dem Alphabet komplexere
Ausdr\"ucke formen kann, wie ``nicht $p$'', ``q oder $r$'', ``nicht(p und
q)'',
``nicht(nicht r)'' etc. (Dies ist die sehr einfache ``Aussagenlogik''.)

Intuitiv: man beschreibt ein festes Objekt, wobei zB ``p'' f\"ur ``rot''
steht, ``q'' f\"ur ``rund'', etc. In der Regel achtet man darauf, dass
die Eigenschaften unabh\"angig voneinander sind, zB ist die Unabh\"angigkeit
mit
``p'' $=$ ``quadratisch'', ``q'' $=$ ``rechteckig'' verletzt, da aus
``quadratisch'' ``rechteckig'' folgt.
\subsubsection{
Herleitung (= Folgerung)
}

Mit Hilfe von Regeln kann man jetzt aus einer Menge von Aussagen
neue Aussagen folgern, zB aus $A=\{p,$ $q$ und $r\}$ kann ich ``q und $r$''
folgern.

Das k\"urzt man ab durch $A \xcl q$ und $r,$ wobei $ \xcl $ das Zeichen
f\"ur ``folgt''
in klassischer Logik ist. Also:
Aus der Menge A folgt in klassischer Logik ``q und $r$''.

Wenn man Folgerungen in anderen Logiken untersucht, benutzt man oft
statt $ \xcl $ das Zeichen $ \xcn.$ $A \xcn q$ und $r$ bedeutet dann:
Aus der Menge A folgt in meiner anderen (mich jetzt interessierenden)
Logik ``q
und $r$''.

Die Regeln und Voraussetzungen werden genau pr\"azisiert, ihnen wird aber
kein besonderer Status wie ``unmittelbar einsichtig'' oder so
zugeschrieben.
\paragraph{
Bemerkungen
}

 \xEh
 \xDH
In vielen Logiken gilt die Regel ``EFQ'' $(=$ ex falso quodlibet), das
heisst,
aus einer widerspr\"uchlichen Aussagenmenge kann man alles Beliebige
folgern,
z. B. auch, dass der Mond aus K\"ase ist. Das ist nichts Tiefsinniges, und
l\"asst sich relativ leicht zeigen, w\"urde hier aber nur verwirren.
 \xDH
Klassische Logiken sind ``monoton'' in dem Sinne, dass, wenn ich die
Menge der Voraussetzungen vergr\"ossere, ich keine m\"oglichen Folgerungen
verliere, kurz, wenn $A \xcc A',$ und $A \xcl a,$ dann auch $A' \xcl a.$
 \xDH
Die obigen Logiken (\"uber Normalf\"alle und das moralisch gebotene) sind
``nicht-monoton'', weil ich durch Vergr\"osserung der Menge der
Voraussetzungen
unter Umst\"anden Folgerungen verliere. Beispiel: Wenn meine einzige
Voraussetzung ``Vogel'' ist, schliesse ich auf ``fliegen'', wenn ich
``Pinguin'' zu den Voraussetzungen hinzunehme, kann ich nicht mehr auf
``fliegen'' schliessen (sondern sogar auf das Gegenteil: ``nicht fliegen'').
Daher findet man das Wort ``nonmonotonic'' im Titel
von z. B.  \cite{Sch18}.
 \xEj
\subsection{
Modelle (= Semantik)
}

Eine Sprache und Logik zu haben, bedeutet noch nicht, zu wissen,
wor\"uber man spricht und schliesst.

Das wird durch den Begriff der Modelle gekl\"art.

In obiger Sprache aus dem Alphabet $p,q,r$ hat man die 8 Modelle, die alle
M\"oglichkeiten beschreiben:
 \xEh
 \xDH $p,q,r$
 \xDH $p,q,$ nicht $r$
 \xDH $p,$ nicht $q,$ $r$
 \xDH $p,$ nicht $q,$ nicht $r$
 \xDH nicht $p,$ $q,r$
 \xDH nicht $p,$ $q,$ nicht $r$
 \xDH nicht $p,$ nicht $q,$ $r$
 \xDH nicht $p,$ nicht $q,$ nicht $r$
 \xEj
das sind die 8 M\"oglichkeiten.

Ein Gegenstand kann rot, rund, zB aus Holz sein, oder eben nicht, alle
8 Varianten sind m\"oglich.

Man kann dann definieren, wann eine Aussage in einem Modell gilt,
und weiter

$A \xcm a$ wenn:
in allen Modellen, in denen alle Aussagen in A gelten, auch die Aussage a
gilt.

Wenn man \"uber andere als klassische Logiken spricht, verwendet man
verwandte Zeichen, z. B. $ \xcm' $ statt $ \xcm.$
\subsubsection{
\"Aquivalenz
}

Ziel logischer Arbeiten ist oft, die \"Aquivalenz von

$ \xcl $ und $ \xcm $ zu zeigen

(f\"ur klassische Logik wurde das nat\"urlich schon
vor langer Zeit gezeigt),

resp. f\"ur eine andere Logik $ \xcn $ die \"Aquivalenz von

$ \xcn $ und einer anders definierten Semantik, $ \xcm'.$

Wenn man das zeigen kann, hat man bewiesen, dass f\"ur diese neue
Logik $ \xcn $ der Herleitungsbegriff dem Modellbegriff
entspricht.

In meinem Buch finden sich viele solche Beweise.

Zusammenfassungen: in Tabellen 1.2 bis 1.6, $S.$ 45-53,
Tabellen 4.1 - 4.5, $S.$ 348-350, aber auch Abschnitt 3.3.2.4, $S.$ 322
ff, etc.
Manchmal kann man aber auch negative Ergebnisse zeigen, z. B., dass
endliche Axiomatisierungen unm\"oglich sind, z. B.
Abschnitt 4.3.2.4, $S.$ 367 ff.

(Alle Angaben beziehen sich auf  \cite{Sch18}.)
\subsection{
Pr\"aferentielle Modelle f\"ur
``Normalit\"at'' und ``Moralit\"at''
}

% {\tiny LABEL: {Section Mini}} \\[1mm]
\label{Section Mini}

(Dieser Absatz ist etwas ``h\"arter''.)

Wenn wir eine Menge von Modellen $M$ haben, und
eine (Pr\"af\"arenz-) Relation $<$ auf $M,$ d.h. f\"ur
einige Modelle $m,m' $ in $M$ gilt $m' <m,$ interessieren wir uns f\"ur die
in $M$
minimalen Modelle, d.h. die $m$ in $M,$ f\"ur die es kein $m' $ in $M$
gibt, mit
$m' <m.$
(Das sind die normalsten oder moralisch besten Modelle in obigen
Interpretationen von $<.)$
 \xEh
 \xDH
Wenn $ \xbm (M)$ die Menge der minimalen Modelle in $M$ ist, gilt
offensichtlich
$ \xbm (M) \xcc M,$ d.h. alle minimalen Modelle in $M$ sind Modelle in
$M.$
 \xDH
Wenn $M' $ eine Teilmenge von $M$ ist, $M' \xcc M,$ und $m$ ist in $M' $
und in $ \xbm (M),$
$m \xbe M' \xcs \xbm (M),$ dann ist $m$ auch in $ \xbm (M').$

Das sieht man am besten so:

Nimm an, $m$ ist in $M' $ und in $ \xbm (M),$ aber nicht in $ \xbm (M'),$
dann muss es
mindestens ein $m' $ in $M' $ geben mit $m' <m.$ Da aber $M' $ eine
Teilmenge von $M$ ist,
ist dieses $m' $ auch in $M,$ also ist $m$ nicht minimal in $M,$
Widerspruch.
Folglich ist (mindestens) eine Annahme falsch, und wenn, wie
vorausgesetzt,
$m$ in $M' $ und in $ \xbm (M)$ ist, muss $m$ in $ \xbm (M')$ sein.
 \xEj

Wir haben also die Eigenschaften
\xEn
 \xDH $(\xbm \xcc)$
$ \xbm (M) \xcc M$
 \xDH $(\xbm PR)$
Wenn $M' \xcc M$ und $m \xbe \xbm (M),$ und $m \xbe M',$ dann auch $m
\xbe \xbm (M'),$ kurz:

Wenn $M' \xcc M,$ dann $ \xbm (M) \xcs M' \xcc \xbm (M')$
\xEp

(Die Namen der Eigenschaften, $(\xbm \xcc)$ und $(\xbm PR),$ erkl\"aren
sich aus
dem Zusammenhang.)

Das Interessante ist, dass diese Eigenschaft (im Wesentlichen)
pr\"aferentielle
Strukturen charakterisiert, d.h. es gilt auch umgekehrt, wenn eine
Auswahlfunktion $ \xbm $ die Eigenschaften
$(\xbm \xcc)$ und $(\xbm PR)$
erf\"ullt, dann gibt es eine Relation $<,$ die genau dieser Auswahlfunktion
entspricht, d.h. f\"ur alle $M$ gilt:

$ \xbm (M)$ $=$ $\{m:$ $m$ ist $<$-minimal in $M\}$

(Siehe Proposition 1.3.1, $S.$ 56.)

Der Autor nennt diesen Teil, d.h. die
Eigenschaften $(\xbm \xcc)$ und $(\xbm PR)$ ``algebraische Semantik'',
weil sie
algebraische Eigenschaften ausdr\"uckt, und die entsprechende Relation $<$
die strukturelle Semantik, weil es eine Struktur ist.

Schliesslich gibt es eine dem entsprechende logische Axiomatik, siehe
Proposition 1.3.20, $S.$ 70. Die entscheidende Eigenschaft ist (PR),
die obigem $(\xbm PR)$ entspricht.

Dies zu erl\"autern w\"urde zu lang werden, und wohl nicht so viel
zus\"atzliche Erkenntnis bringen. Trotzdem ein Beispiel der Eigenschaften
dort:

(LLE) $ \ol{T}= \ol{T' }$ $ \xcp $ $ \ol{ \ol{T} }= \ol{ \ol{T' } }$

bedeutet (in meiner Notation): wenn die klassischen Folgerungen von $T$
und $T' $
gleich sind, dann auch die Folgerungen in der Logik $ \xcn.$ (Dies ist
trivial,
wenn die Logik \"uber Modellmengen definiert ist.)

Wir haben also die \"Aquivalenzen:

Axiomatik $ \xcj $ Algebraische Semantik $ \xcj $ Strukturelle Semantik

(Diese Aufteilung charakterisiert auch viele meiner Arbeiten, sie hat sich
als
sehr fruchtbar herausgestellt, statt direkt

Axiomatik $ \xcj $ Strukturelle Semantik

zu zeigen, da man die verschiedenen Probleme so besser trennen kann.)
\section{
Maschinelles Lernen, neuronale Netze
}

Maschinelles Lernen, wor\"uber heute viel gesprochen wird,
basiert auf einer
Selbst-Organisation von ``neuronalen Netzen''. Der Autor weiss dar\"uber fast
nichts, nur dass das
menschliche Gehirn wohl st\"arker strukturiert ist als neuronale Netze,
aufgeteilt in verschiedene Gehirn-Regionen mit zum Teil unterschiedlicher
interner Struktur und komplexen Verbindungen untereinander.
\section{
R\"uckblick
}

Der oben skizzierte
logische Ansatz hat sicher zu Kl\"arung von philosophischen und
mathematischen Aspekten gef\"uhrt, vielleicht weniger zur KI beigetragen,
ich habe ihn immer auch als ``experimentelle Philosophie'' betrachtet.
\clearpage
\section{
Beitr\"age des Autors
}
% ++

Ich habe meist versucht,
 \xEh
 \xDH neue Konzepte zu entwickeln,
 \xDH neue Beweismethoden zu entwickeln,
 \xDH etwas schwierigere Beweise zu f\"uhren.
 \xEj

``Kap.'' bezieht sich - wenn nicht anders explizit gesagt - auf die Kapitel
der
beiden B\"ande  \cite{Sch18}.
\subsection{
Konzepte
}

An mehr oder weniger neuen Konzepten habe ich untersucht:
 \xEh
 \xDH algebraische vs. strukturelle Semantik (Kap. 1.2),
 \xDH abstrakte Gr\"osse als algebraische Semantik f\"ur nicht-montone
Logiken,
in Aussagen- und Pr\"adikatenlogik,
sowie additive und multiplikative Eigenschaften davon, sowie ihre
Verbindung zu logischen Eigenschaften, u.a. auch zu (logischer)
Interpolation.
 \xDH abstrakte Distanz als algebraische Semantik f\"ur Theorie Revision,
Update, Deontische Logik (Kap. 4.3),
 \xDH Definierbarkeits-Erhaltung (definability preservation), d.h.,
ob Operatoren definierbare Modellmengen wieder in definierbare
Modellmengen \"uberf\"uhren, und was passiert, wenn nicht (Kap. 1.7),
 \xDH Abgeschlossenheit der Modellmengen unter $ \xcs $ und $ \xcv $ (Kap.
1.5),
 \xDH Homogenit\"at - das Problem wurde in
 \cite{Sch97-2}, Kapitel 1.3.11, angesprochen, aber erst
in  \cite{GS16}, Kapitel 11, bearbeitet, siehe hier
Kap. 5.7.
Merkw\"urdigerweise wird das Problem in der KI allgemein kaum beachtet, was
daran
liegen mag, dass es auch ein Problem der Erkenntnistheorie ist.
 \xEj
\subsection{
Methoden
}

 \xEh
 \xDH
Meine wichtigste, und fast immer angewandte Methode, ist,
Vollst\"andigkeitsbeweise in zwei Unterbeweise zu trennen, einmal zwischen
Axiomatik und algebraischer Semantik, und dann zwischen algebraischer und
struktureller Semantik. Das trennt die Probleme viel klarer, als Beweise
in einem Schritt zu f\"uhren, und zeigt gemeinsame Probleme (zum Beispiel
mit Definierbarkeit) in teils sehr verschiedenen Beweisen auf.

Z. B. hat die Existenz von interpolierenden Formeln zum Teil mit der
Aussdrucksst\"arke der jeweiligen Sprache zu tun, dass interpolierence
Modellmengen zwar existieren, aber nicht beschrieben werden k\"onnen. Das,
und der Zusammenhang zwischen multiplikativen Eigenschaften der abstrakten
Semantik und Existenz von interpolierenden Modellmengen war zB unerwartet
(Kap. 6).
 \xDH
Mehr im Detail ist meine Methode, Modelle in pr\"aferentiellen
Strukturen mit geeigneten Auswahlfunktionen, oder, in komplizierteren,
transitiven, F\"allen, mit geeigneten B\"aumen, zu indizieren, sehr
erfolgreich,
sie hat auch erlaubt, Beweismethoden auf neue Bereiche zu \"ubertragen,
zB auf Mengen von Folgen (Kap. 1.3).
 \xDH
Zum Beweis der Unm\"oglichkeit von Axiomatisierungen gewisser Gr\"osse
(endlich, oder unendlich, aber von fester unendlicher Gr\"osse)
habe ich, wie \"ublich, Diagonalmethoden benutzt, sie dem jeweiligen
Fall angepasst (u.a. ``Hamsterrad'' in der Theorie-Revision, und die weit
komplizierteren Beweise f\"ur pr\"aferentielle Strukturen ohne
Definierbarkeits-Erhaltung resp. die $ \xbl $-Variante hiervon.)
(Kap. 1.7, 4.3.2)
 \xEj
\subsection{
Resultate
}

Meine wichtigsten Resultate sind wohl
 \xEh
 \xDH eine Reihe von Repr\"asentationss\"atze f\"ur
pr\"aferentielle Logiken,
auch die Trivialisierung resp. Unl\"osbarkeit im \"ublichen Sinne
der $ \xbl $-Variante (Kap. 1.3-1.7),
 \xDH die generelle Bedeutung von Definierbarkeits-Erhalt, sowie von
Abschlusseigenschaften der Definitionsmenge (Kap. 1.7),
 \xDH Repr\"asentation von ``normalerweise'' in der Pr\"adikatenlogik (Kap.
3.3),
 \xDH Distanz-Semantik f\"ur Theorie Revision (mit Lehmann und Magidor),
Unm\"oglichkeit endlicher Axiomatisierung (Kap. 4.3),
 \xDH Interpolation und Verbindung zu multiplikativen Eigenschaften
abstrakter Semantik (Kap. 6.5),
 \xDH abstrakte Unabh\"angigkeit von Funktionenprodukten (Kap. 8).
 \xEj

\end{document}